\documentclass[aps,twocolumn,amsmath,amssymb,superscriptaddress, nofootinbib, nobibnotes, showkeys, 12pt]{revtex4-1}

\usepackage{tabularx}
\usepackage[table,xcdraw]{xcolor}
\usepackage{booktabs}
\usepackage{graphicx}
\usepackage{dcolumn}
\usepackage{bm}
\usepackage{url}
\usepackage{multirow}
\usepackage[colorlinks=true,linkcolor=blue,citecolor=blue,urlcolor=blue]{hyperref}

\usepackage{xr}
\begin{document}
\pagecolor{white}
\title{Automated Segmentation and Analysis of Microscopy Images of Laser Powder Bed Fusion Melt Tracks}

\begin{abstract} 
With the increasing adoption of metal additive manufacturing (AM), researchers and practitioners are turning to data-driven approaches to optimise printing conditions. Cross-sectional images of melt tracks provide valuable information for tuning process parameters, developing parameter scaling data, and identifying defects.  Here we present an image segmentation neural network that automatically identifies and measures melt track dimensions from a cross-section image.  We use a U-Net architecture to train on a data set of 62 pre-labelled images obtained from different labs, machines, and materials coupled with image augmentation. When neural network hyperparameters such as batch size and learning rate are properly tuned, the learned model shows an accuracy for classification of over 99\% and an F1 score over 90\%.  The neural network exhibits robustness when tested on images captured by various users, printed on different machines, and acquired using different microscopes. A post-processing module extracts the height and width of the melt pool, and the wetting angles. We discuss opportunities to improve model performance and avenues for transfer learning, such as extension to other AM processes such as directed energy deposition.
\end{abstract}

\keywords{laser powder bed fusion, machine learning, metal additive manufacturing}

\author{Aagam Shah}
\affiliation{Department of Materials Science \& Engineering, 1304 W Green Street, University of Illinois at Urbana-Champaign, Urbana IL 61801}
\author{Reimar Weissbach}
\affiliation{Department of Mechanical Engineering, Massachusetts Institute of Technology, 77 Massachusetts Avenue, Cambridge, MA, 02139, USA}
\author{David A. Griggs}
\affiliation{Department of Mechanical Engineering, Massachusetts Institute of Technology, 77 Massachusetts Avenue, Cambridge, MA, 01239, USA}
\author{A. John Hart}
\email[Corresponding author. Tel: 617 324-7022. E-mail: ]{ajhart@mit.edu}
\affiliation{Department of Mechanical Engineering, Massachusetts Institute of Technology, 77 Massachusetts Avenue, Cambridge, 02139, MA, USA}
\author{Elif Ertekin}
\email[Corresponding author. Tel: 217 333-8175. E-mail: ]{ertekin@illinois.edu}
\affiliation{Department of Mechanical Science \& Engineering, 1206 W Green Street, University of Illinois at Urbana-Champaign, Urbana IL 61801}
\author{Sameh Tawfick}
\email[Corresponding author. Tel: 217 244-6303. E-mail: ]{tawfick@illinois.edu}
\affiliation{Department of Mechanical Science \& Engineering, 1206 W Green Street, University of Illinois at Urbana-Champaign, Urbana IL 61801}
\affiliation{Materials Research Laboratory, University of Illinois at Urbana-Champaign, Urbana IL 61801}
\date{\today}
\maketitle

\section{Introduction}

Additive manufacturing (AM) by laser powder bed fusion (LPBF) can create near net shape components with intricate geometries, without specialized tooling, and finds commercial applications in aviation components, medical implants, and bespoke mechanical systems, among others~\cite{adelmann2022mechanical, lantada2024additive, garcia2020comparison, madhavadas2022review}. 
One of the most promising avenues in LPBF is the development and application of new, tailored materials, including advanced designs such as multi-material gradient structures~\cite{demir2022enabling, gunther2018design, li2019progress}. 

LPBF practitioners that want to optimize processing strategies (e.g., to improve production rate or the quality of parts) or utilize new materials have to develop suitable processing parameters. This is often pursued via iterative trial and error experimentation. Laser power, scan speed, laser spot size, hatch spacing, and powder layer thickness are among the parameters that have to be optimized for a successful build. The initial parameter space is often very large, requiring a large number of experiments and analysis.

While physics-based computational models are not yet sufficiently developed and available at scale, data-driven optimisation methods are a promising avenue to support parameter development~\cite{park2024data, cao2021optimization, parsazadeh2023new}. To generate a sufficient amount of data for this purpose, single layer and single track experiments are significantly more efficient to perform than full builds when doing parameter development studies. 
Exemplary research directions informed by single-layer and single-track studies are powder spreading~\cite{oropeza2022mechanized, chen2022high, Penny2021, PENNY2024Blade, PENNY2024Roller, weissbach2024exploration}, melt pool dynamics and scaling~\cite{nayak2020effect, nayak2021effect, martucci2021automatic, makoana2018characterization, matthews2017denudation, simson2024experimental, li2012balling, chen2022high, kouprianoff2017line, yuan2022understanding, zhang2020experimental, yadroitsev2010selective, nie2023effect, hanemann2020dimensionless, weaver2022laser, lane2020measurements}, new material development~\cite{ghasri2023single}, microstructure evolution~\cite{mohammadpour2022microstructure, nie2023effect}, and formation of porous materials in LPBF~\cite{jafari2020porous}.

To interpret results from single-track experiments, manual analysis of optical microscope images of the melt track cross-sections is typically performed. These images provide quantitative information that can be used to understand the manufacturing regime, such as the height and width of the melt pool, as well as wetting angles, which makes the analysis a critical step in generating research results. Manual image analysis is time consuming and can be prone to human error. 
Moreover, data-driven approaches to interpret LPBF, such as through scaling relationships among dimensionless process parameters, requires a large number of cross-sectional images~\cite{Fabbro2019Scaling, Hanemann2020Dimensionlessa, Hann2011Simplea, King2014Observationa, Naderi2023Fidelitya, Rubenchik2018Scaling, Ye2019Energya}. 
Accordingly, in recent years, machine learning and deep learning techniques have been employed to simplify the task of analysing different types of images of metal AM techniques~\cite{Baumgartl2020Deep, Fang2021Insitu, Jardon2022Effect, Liu2021Melt, Schmid2021New, Yeung2020Meltpool, Ogeke2024Deep, Krishna2024Deep}. 
Many of these techniques have been developed for LPBF~\cite{Schmid2021New, Yeung2020Meltpool, Baumgartl2020Deep}, but none tackle the single-track cross-sectional images. 

In this work, we demonstrate the application of U-Net to automated segmentation of optical microscopy images of cross-sections of melt tracks from LPBF experiments. Our strategy used images of single melt track obtained from two research labs using different machines, sample preparation approaches, and microscopes to build a versatile training data set. We use software developed in-house to first generate a reliable data set of segmented training images. Using this data set, we train a U-Net neural network to segment cross-sectional images of LPBF melt tracks into the region of the melt track and the surrounding region, and show that it can automatically and rapidly quantify the height, depth, width, wetting angles, and wall angles of a melt pool. The model trained on only single track images is additionally able to segment some multi-track images with reasonable accuracy, showing promise for transfer learning.
This work presents an automated technique for the analysis of optical microscopy images of cross-sections of LPBF melt tracks, towards fostering closed loop optimisation of LPBF.
The trained model is available for general use, and provided with an installable software for ease of use~\cite{shah2024software}.

\section{Methods}

\subsection{Experiments and Imaging}

\begin{figure*}
	\centering
	\includegraphics[width = 0.99\linewidth]{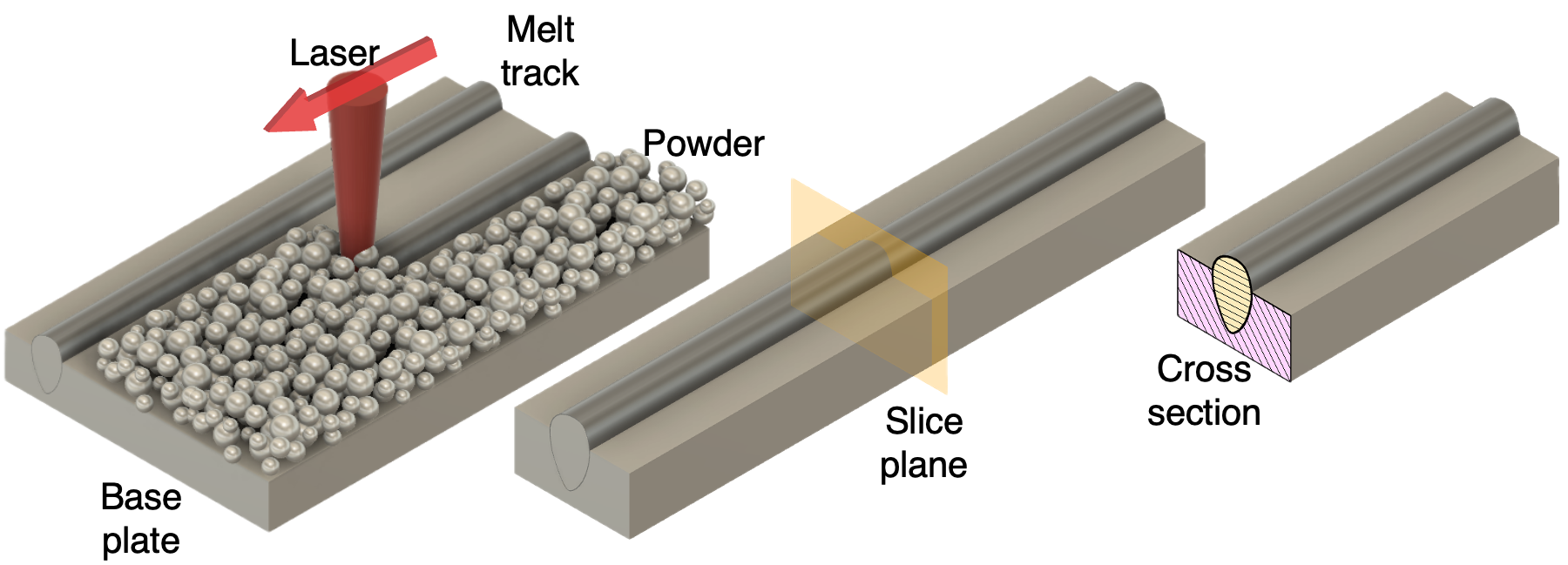}
	\caption{\label{fig:experiment-schematic}Schematic of the experimental process flow from single layer melting to cross-sectioning. Many parallel single-tracks are typically melted in a single experiment, and subsequently cross-sectioned, and then imaged separately from the sample. The schematic shows sectioning of only single tracks.}
\end{figure*}

\begin{figure*}
	\centering
	\includegraphics[width = 0.99\linewidth]{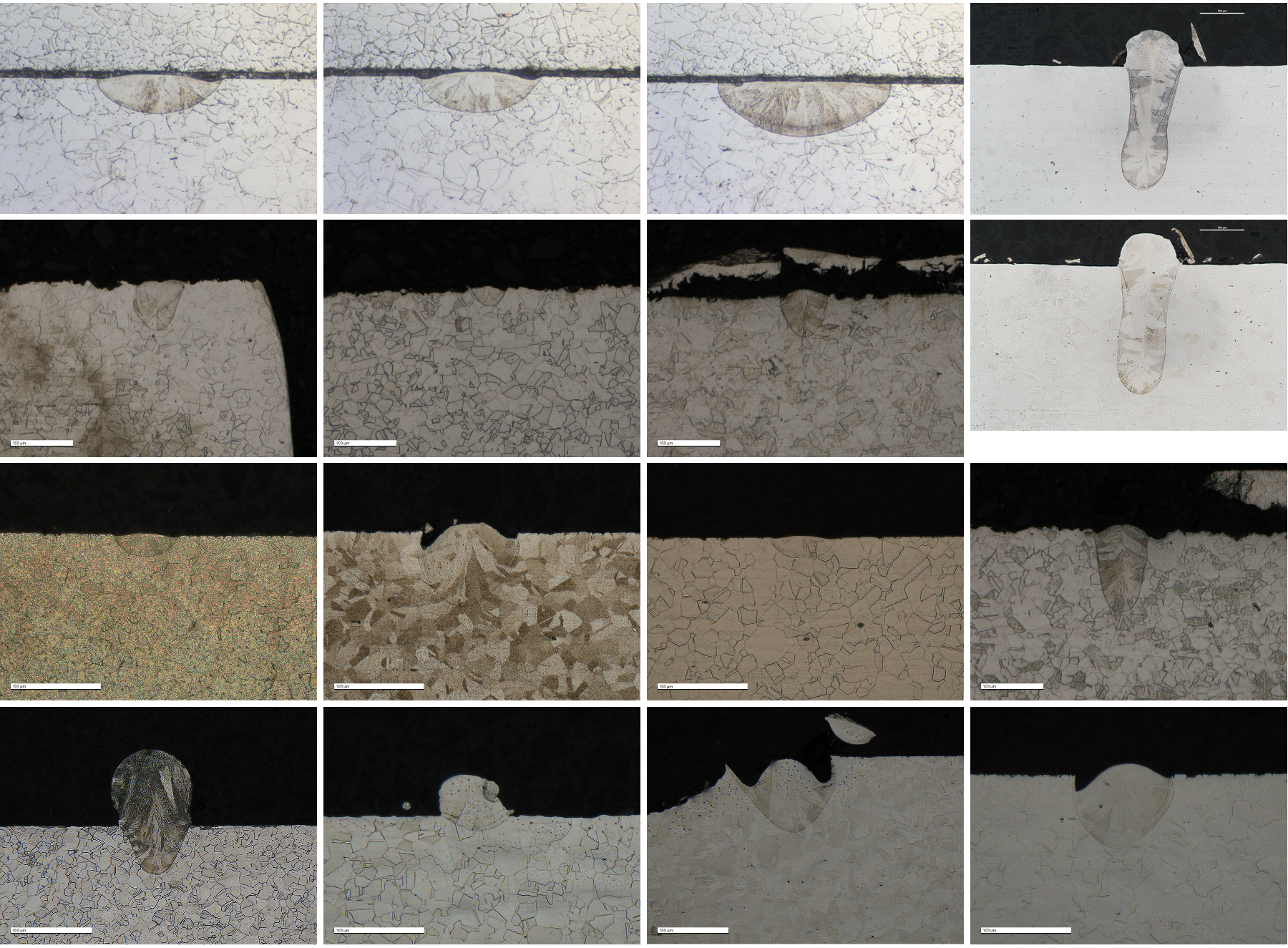}
	\caption{\label{fig:data-diversity} Example micrographs of melt track cross-sections showing a diversity in the contrast, material, colour profile, and aspect ratio. The images present artefacts such as scale bars of different types and in various locations, excess powder, and an additional plate.}
\end{figure*}

\begin{figure}
	\centering
	\includegraphics[width = 0.99\linewidth]{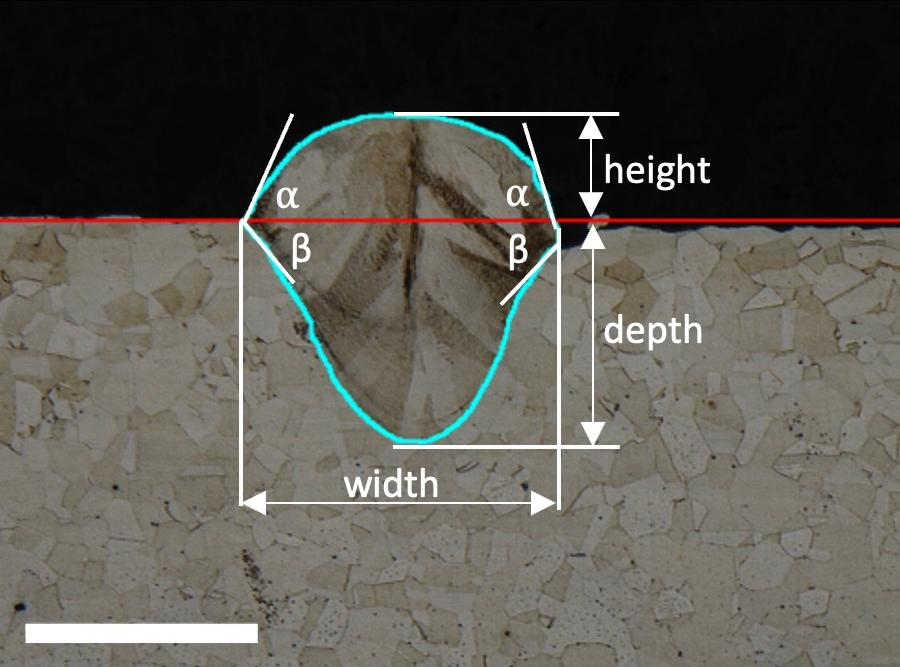}
	\caption{\label{fig:metrics-schematic}Ideal melt track cross-section showing the extracted geometrical metrics: melt track depth, height, width, wetting angle $\alpha$, wall angle $\beta$. The red line shows the baseline for the substrate surface and cyan line shows the border of the melt pool. The scale bar is 100 $\mu$m. }
\end{figure}

The employed neural network architecture was trained on a curated data set of cross-sectional images of single line melting experiments.  Images from three sources were used: bare plate melting experiments performed using a bespoke miniature LPBF testbed~\cite{griggs2022testbed}; melt track cross-sections produced via the National Institute for Standards and Technology (NIST) Additive Manufacturing Metrology Testbed (AMMT)~\cite{lane2016design}; and single-layer melting experiments performed on a commercial LPBF machine (EOS M290).

For the bare plate experiments, we utilized SS316 substrates (0.25" thick) that were sandblasted on the top for improved laser absorptivity. Laser scan lines had a 1~mm spacing and were printed perpendicular to the gas flow direction. The laser spot size was held constant at $4 \sigma$~=~109~$\mu m$, and the scan speed at 1~m/s, with power settings of 60~W, 130~W, and 200~W, as published in~\cite{griggs2022testbed}. 

The second set of training images derived from bare plate experiments was provided by NIST and is available publicly ~\cite{lane2016design, weaver2022laser, lane2020measurements}. The experiments employed nickel superalloy 625 (IN625) substrates, and a wide range of laser process parameters~\cite{weaver2022laser}.

Third, single powder layer melting experiments were performed using a commercial EOS M290 LPBF machine under standard operating conditions. High-precision etched SS316 single layer templates were used a high level of control over powder layer consistency. 15-45~$\mu m$ SS316L powder was manually spread with a machinist blade of 1/8 in thickness into templates with depths ranging from 60~$\mu m$ to 150~$\mu m$. The powder size distribution is D10~=~22.6$\mu m$, D50~=~36.1$\mu m$, D90~=~56.0$\mu m$ (Carpenter Technology). For more detail on the experimental method and single layer template preparation, the reader is referred to ~\cite{weissbach2024workflow}.

After laser melting, the post-processing of the melt tracks is similar across all three sources of experimental data, as schematically shown in Figure~\ref{fig:experiment-schematic}. The substrate plates were cut perpendicular to the melt track direction with a low impact cutting method such as a diamond blade saw, then ground, polished, and subsequently etched with Aqua Regia (1pt water, 1pt nitric acid, 1pt hydrochloric acid) to reveal the metallographic microstructure and solidified melt pool boundaries.

Figure~\ref{fig:data-diversity} shows exemplary images from our training data, with these examples being selected to highlight the diversity of the data, for instance: shallow vs. deep melt pools; a bare plate without new material being deposited vs. significant material being melted to the substrate; different levels of etching, color, and contrast. From each of the cross-sections, geometrical features include the melt track depth, the amount of the melt track above the substrate surface line (i.e., new material being deposited) vs. below the substrate surface line, melt track width, cross-sectional area, wetting angle $\alpha$, as well as wall angle $\beta$ (see Figure~\ref{fig:metrics-schematic}).

\subsection{Generation of Training Data}

\begin{figure*}
	\centering
	\includegraphics[width = 0.99\linewidth]{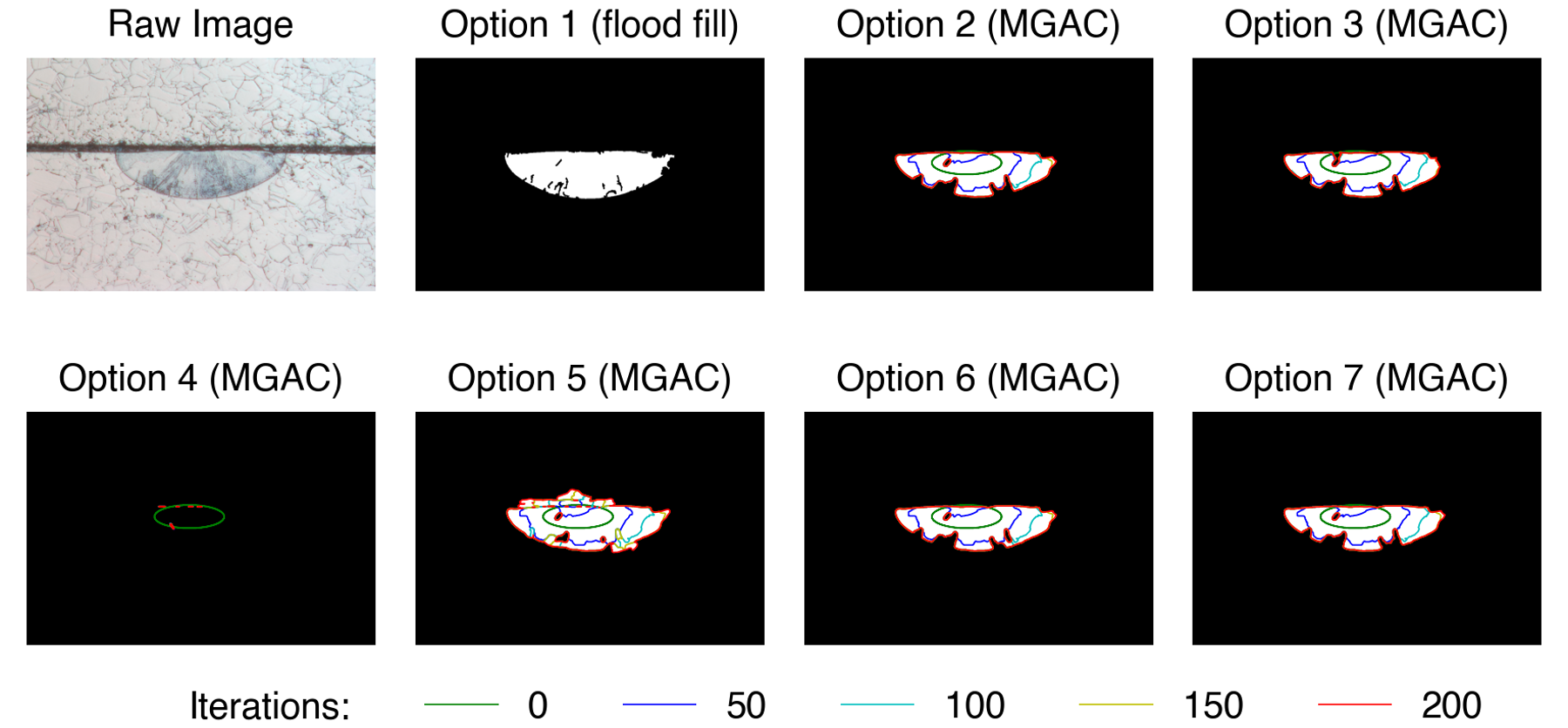} \\[0.1cm]
        \caption{\label{fig:mgac} Output from the software that uses morphological geodesic active contours (MGAC) to identify the melt pool boundary. The first figure is the raw image. The second figure is a flood fill of the central region after edge detection. Each of the remaining output figures correspond to a different set of hyperparameters for the MGAC based technique. Within each MGAC Option, a small number of the ballooning iterations is depicted using the colors in the legend at the bottom of the figure.}
\end{figure*}

To speed up and standardize the manual segmentation of the large number of images needed for the training, we developed a software that uses morphological geodesic active contours (MGAC)~\cite{MarquezNeila2014Morphological} as the first step to generate masks for our images. The user sets an initial small diameter ellipse as a nucleus within the melt track to initiate the process. Based on an ``energy field" defined by edge detection and subsequent blurring, the initial ellipse is iteratively ballooned, growing its size and conforming to the contour edge of the melt track from the inside out. The software outputs 7 different masks, each corresponding to a different set of hyperparameters. A user may then select the most visually accurate mask. Fig. \ref{fig:mgac} shows an example image and its corresponding masks generated by the active contours method. 

\begin{figure*}
    \includegraphics[width = 0.99\linewidth]{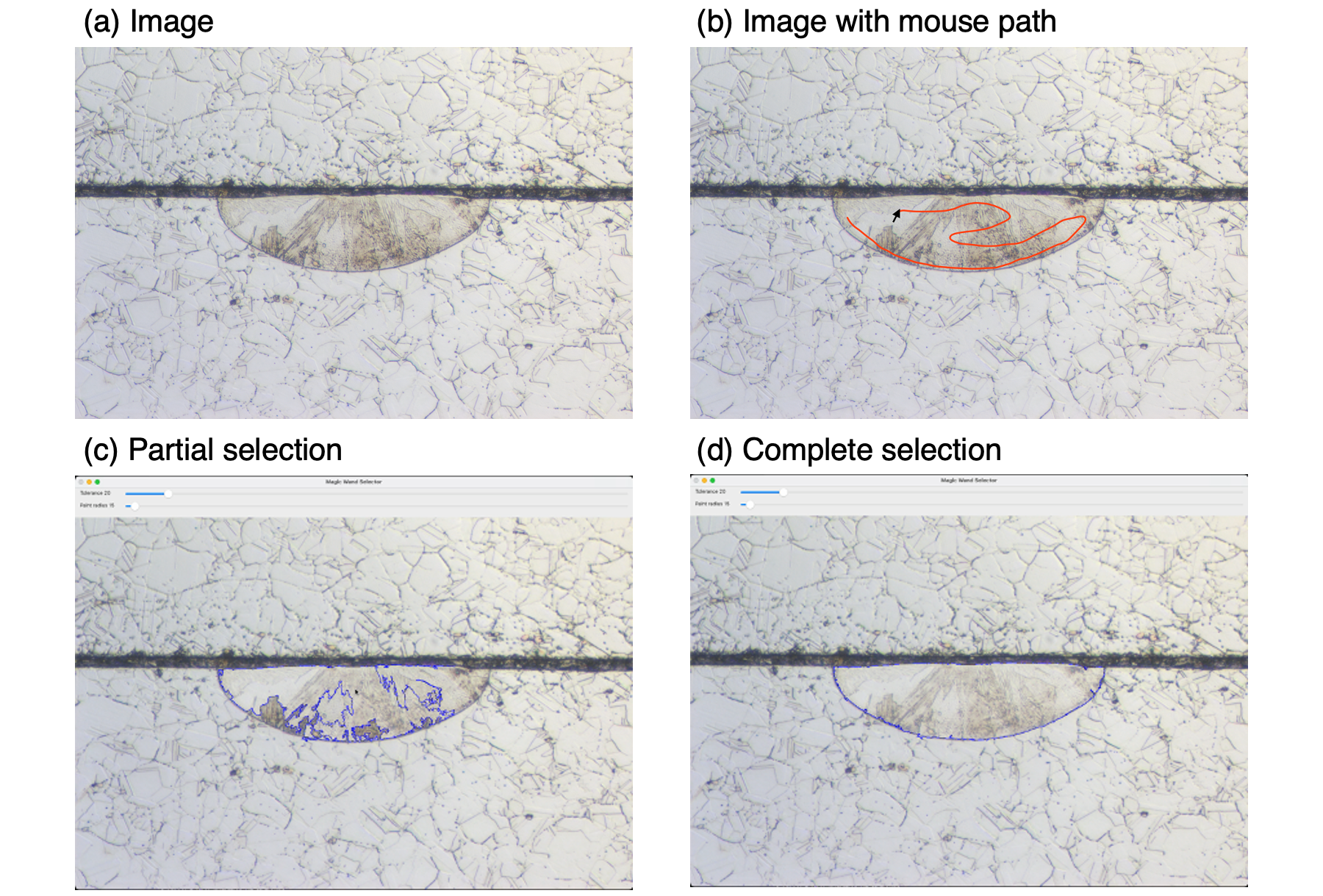}
	\caption{\label{fig:magicwand} Steps while using connected colour thresholding~\cite{Kulkarni2012Color} to segment an image. (a) The raw image; (b) a schematic of the mouse path taken to select the whole region; (c) a snapshot of the region selected halfway through the mouse path in (b); (d) the complete selection. }
\end{figure*}

For cases where none of these segmentation masks are satisfactory to the user to include in the training data set, we built a second software based on connected colour thresholding, which relies more on manual human intervention. The user can paint over a region in the image, for example within the melt track in this case, and the software selects all connected pixels with colour intensity values within a user-defined threshold of the selected pixels. Fig \ref{fig:magicwand} shows the different steps of using this tool with an example image. This is a more labour intensive method, but produces very accurate results because the melt pool boundaries have a colour that is distinct enough from the centre of the melt pool and from grain boundaries. A simple flood fill takes care of all dislocations withing the melt pool. Both these tools are available as an installable python package and software~\cite{shah2024software}.

\begin{table}[]
\begin{tabular}{ccc}
\hline
Operation & \, & Range \\ \hline
Rotation    & & $\pm 20^{\circ}$     \\ 
Width Shift & & 5\%                  \\ 
Height Shift & & 5\%                  \\ 
Shear       & & $0.05$ radians       \\ 
Zoom        & & $95-105$\%           \\ 
Brightness  & & $\gamma = [0.2,1.8]$ \\ 
Vertical Flip  & & 50\% \\ 
Horizontal Flip  & & 50\% \\ \hline
\end{tabular}
\caption{\label{table:augs} Parameters used for image augmentation.}
\end{table}

Using these tools, we generated 76 images, split into 62 training, 6 validation, and 8 testing images. To capitalise on the relatively small data set of images, we employed data augmentation to generate and provide images of varying shape, size, contrast, melt track location, and other sources of complexity in cross-sectional images of melt tracks. For each augmented batch, an image was selected and the following augmentations were performed: rotation, shift, shearing, magnification, horizontal and vertical flipping, and brightness modification. The parameters for these augmentations are summarized in Table \ref{table:augs}. For each image from the initial dataset, we generated 15 augmented images, which gave us a complete training dataset of 992 images. 

\subsection{U-Net Neural Network}

\begin{figure}
	\centering
	\includegraphics[width = 0.99\linewidth]{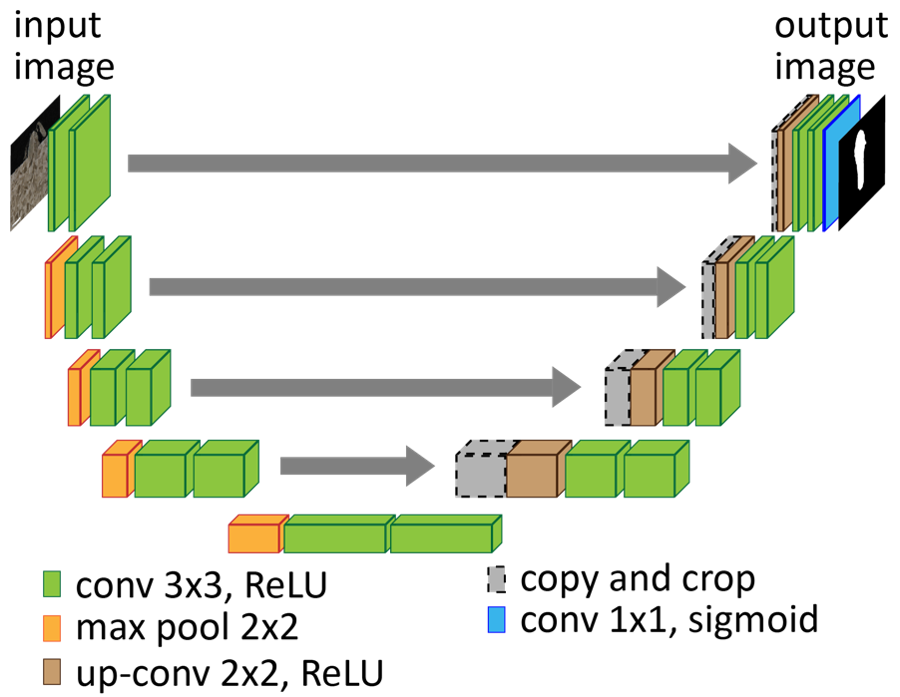}
	\caption{\label{fig:unet-architecture} The U-Net architecture. The encoding half of the U-Net reduces the input to a latent space representation where the relevant pixels are activated and the decoding half brings the latent space representation back to 2D space with the copy and crop steps imparting contextual and spacial information (adapted from~\cite{shah2023automated}).}
\end{figure}

U-Net~\cite{ronneberger2015u} is a convolutional neural network (CNN) architecture that reduces an input image to a latent space representation with successive downsampling. This latent representation is then upsampled with each layer including additional input from the downsampling steps to impart contextual information to the output. To create the neural network, we used the Keras/Tensorflow python framework run in parallel across 8 Volta100 GPUs.

The specific U-Net architecture used in this work is illustrated in Fig. \ref{fig:unet-architecture}. The input image (512 × 512 resolution) is first run through two convolution layers (with padding) with 3 × 3 tile sizes. The output is then put through a 2 × 2 max pooling layer. This process is repeated three more times. The first convolution increases the number of channels to 64 and for each subsequent level, the number of channels is doubled until there are 1024 channels. Each upsampling step goes through a 3 × 3 convolution layer that halves the number of channels. It is then concatenated with the output from the convolution layer from a corresponding downsampling step, as indicated by the gray arrows in Fig. \ref{fig:unet-architecture}. Following two more convolution layers, the number of channels is halved. The upsampling step is repeated three more times until the number of channels is again 64. The output is then fed into one last 3 × 3 convolution layer with two input channels and a 1 × 1 convolution layer with a single output channel. With the exception of the last layer which has a sigmoidal activation, all activations are rectified linear unit (ReLU) functions. The model was trained with batch sizes of 8, 16, and 32 images for 100 epochs. We used the binary cross entropy to calculate the loss as:
\begin{equation}
    L(x,z) = -(z\log(g(x)) + (1 - z)\log(1 - g(x)))
\end{equation}
where $x$ are the logits (converted from the output layer), $z$ are the labels (i.e. 0 or 1), and $g$ is the sigmoid function. To obtain gradient updates, we used the RMSprop optimizer with an initial learning rate of 0.0001. 

\subsection{Post-Processing}
From the binary segmented image, we extract the true surface and baseline first. If the intensity of the top half of the image is too high, we change it to have zero intensity; this tackles the case where there is another sample in the top half of the image. We then identify the top edge using a Gaussian blur followed by a sobel filter. This is the true surface of the metal. To identify the baseline (refer to Fig. \ref{fig:metrics-schematic}), we fit a line to the points in the vicinity of the melt pool.

Once we have the baseline, we identify the bounding box of the melt track. The height and depth of the melt track are calculated with respect to the baseline, and the width is the same as the bounding box. 

To calculate the wetting and wall angles, we identify the point where the true surface intersects the melt track as the tangent point. There is one such point on each side of the melt track. We then identify the tangents to the melt pool at those points in the top and bottom directions, and calculate their angle with the baseline. We report the mean of the wetting and wall angles on the left and right sides.

\section{Results and Discussion}

\subsection{Hyperparameter Optimisation}

\begin{figure}
	\centering
	\includegraphics[width = 0.99\linewidth]{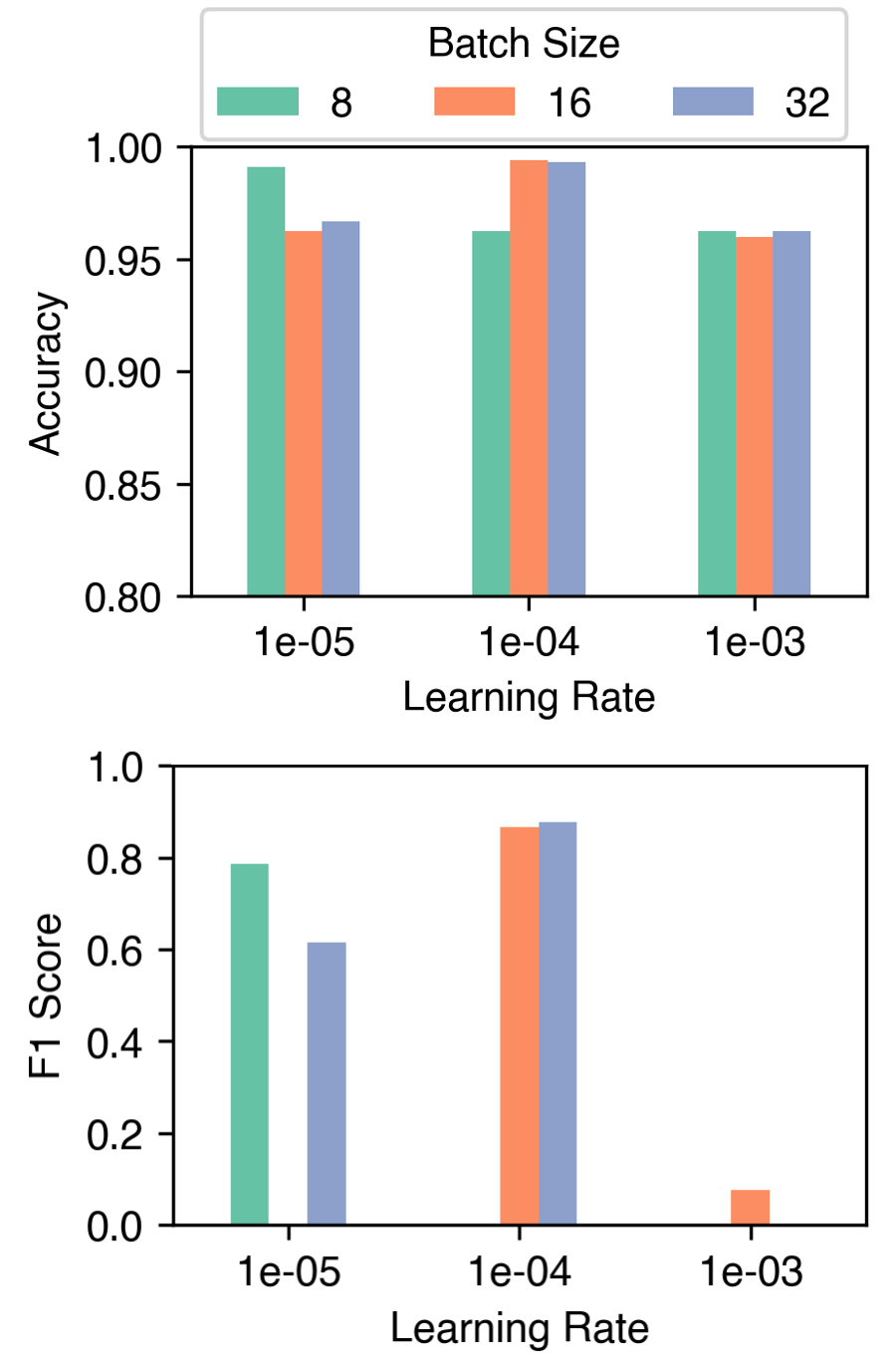}
	\caption{\label{fig:hyperparameter-optimisation} Mean accuracy and F1 score for models trained with different combinations of learning rates and batch sizes all measured on the validation data set. A learning rate of 10$^{-4}$ and a batch size of 16 gave the best results. Missing bars indicate zeros, i.e. the loss function could not be minimised.}
\end{figure}

We investigated the effect of hyperparameters on the model performance, specifically batch size and learning rate.
The metrics used to evaluate the model are the accuracy and the F1 score: 
\begin{align}
    {\rm Accuracy} &= \frac{TP+TN}{TP+TN+FP+FN} \hspace{0.5em} \\
    {\rm F1\,\,Score} &= \frac{2 TP}{2 TP+FP+FN} \hspace{0.5em}
\end{align}
where $TP$ is the number of true positives (pixels correctly labeled as part of the melt pool), $TN$ is the number of true negatives (pixels correctly labeled as not part of the melt pool), $FP$ is the number of false positives (pixels incorrectly labeled as part of the melt pool), and $FN$ is the number of false negatives (pixels incorrectly labeled as not part of the melt pool). 
Intuitively, accuracy is the percentage of pixels correctly classified. As the harmonic mean of the recall (the proportion of true melt pool pixels that are correctly classified) and the precision (proportion of the predicted melt pool pixels that are correctly classified), the F1 score is a measure of the similarity between the prediction and the ground truth. 
Figure \ref{fig:hyperparameter-optimisation} shows the best values obtained for mean accuracy and mean F1 score when training with a batch size of 8, 16 and 32, and an initial learning rate of $10^{-5}$, $10^{-4}$ and $10^{-3}$.
A batch size of 16 and an initial learning rate of $10^{-4}$ gave the best results. 

\subsection{Model Performance}

\begin{figure}
	\centering
	\includegraphics[width = 0.99\linewidth]{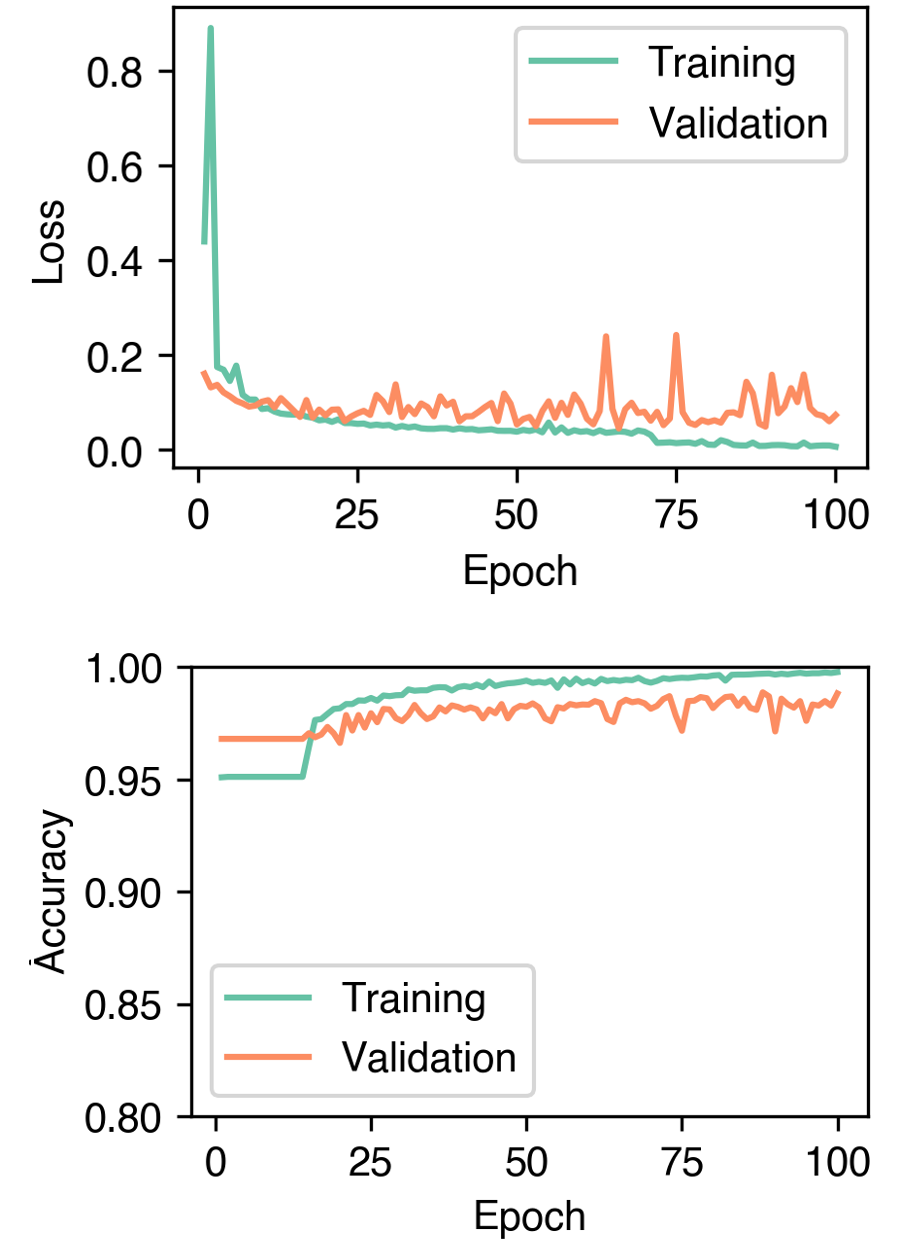}
	\caption{\label{fig:model-training} Training and validation loss and accuracy vs. epoch.  }
\end{figure}

The trained model performs well on most cross-sectional images of LPBF melt tracks. It is capable of accurately identifying melt pools in images having different contrasts and colour profiles, images of different materials, and images of tracks made on different machines.
As shown in Figure \ref{fig:model-training}, both the training and validation losses decrease monotonically with training, and the difference between them is small, indicating that the model does not suffer from overfitting.
The trained model has an accuracy of 0.995, an F1 score of 0.904, and an intersection over union (IoU) of 0.868. IoU is a common metric used to assess cases where a single domain must be identified and is defined as
\begin{align}
    {\rm IoU} &= \frac{TP}{TP+FP+FN} \hspace{0.5em}
\end{align}
Importantly, such high values could be achieved with relatively small training data due to the diversity and high quality of data used~\cite{shah2023automated}. 

\begin{figure}
	\centering
	\includegraphics[width = 0.99\linewidth]{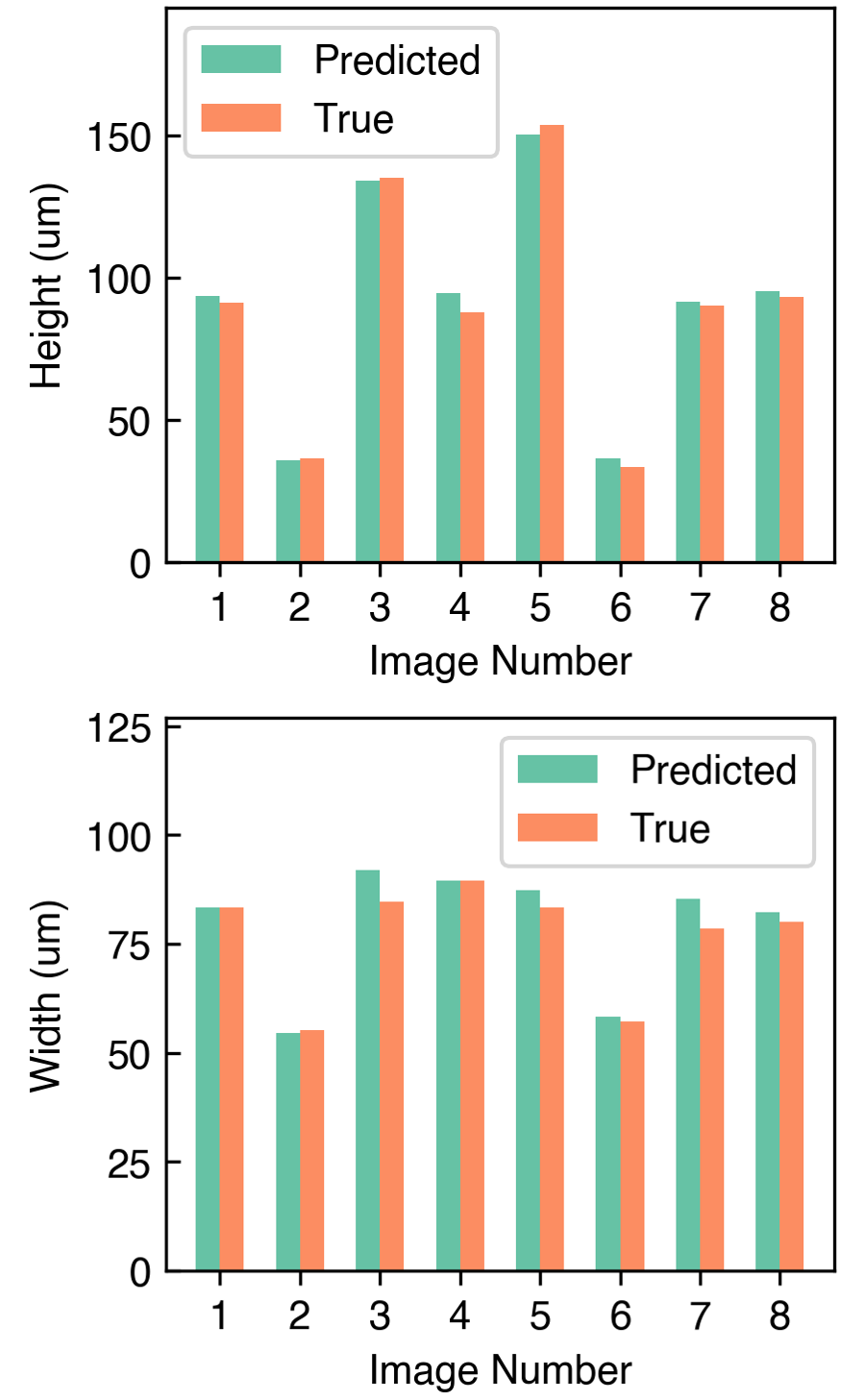}
	\caption{\label{fig:height-width-accuracy} True and predicted values for the height of the entire melt pool for 8 new images. The predicted values agree well with the true values. }
\end{figure}

We show the performance of the model on 8 unseen images in Fig. \ref{fig:height-width-accuracy}. Overall, the predictions are in good agreement with the true values for both height and width.

\begin{figure*}
	\centering
	\includegraphics[width = 0.99\linewidth]{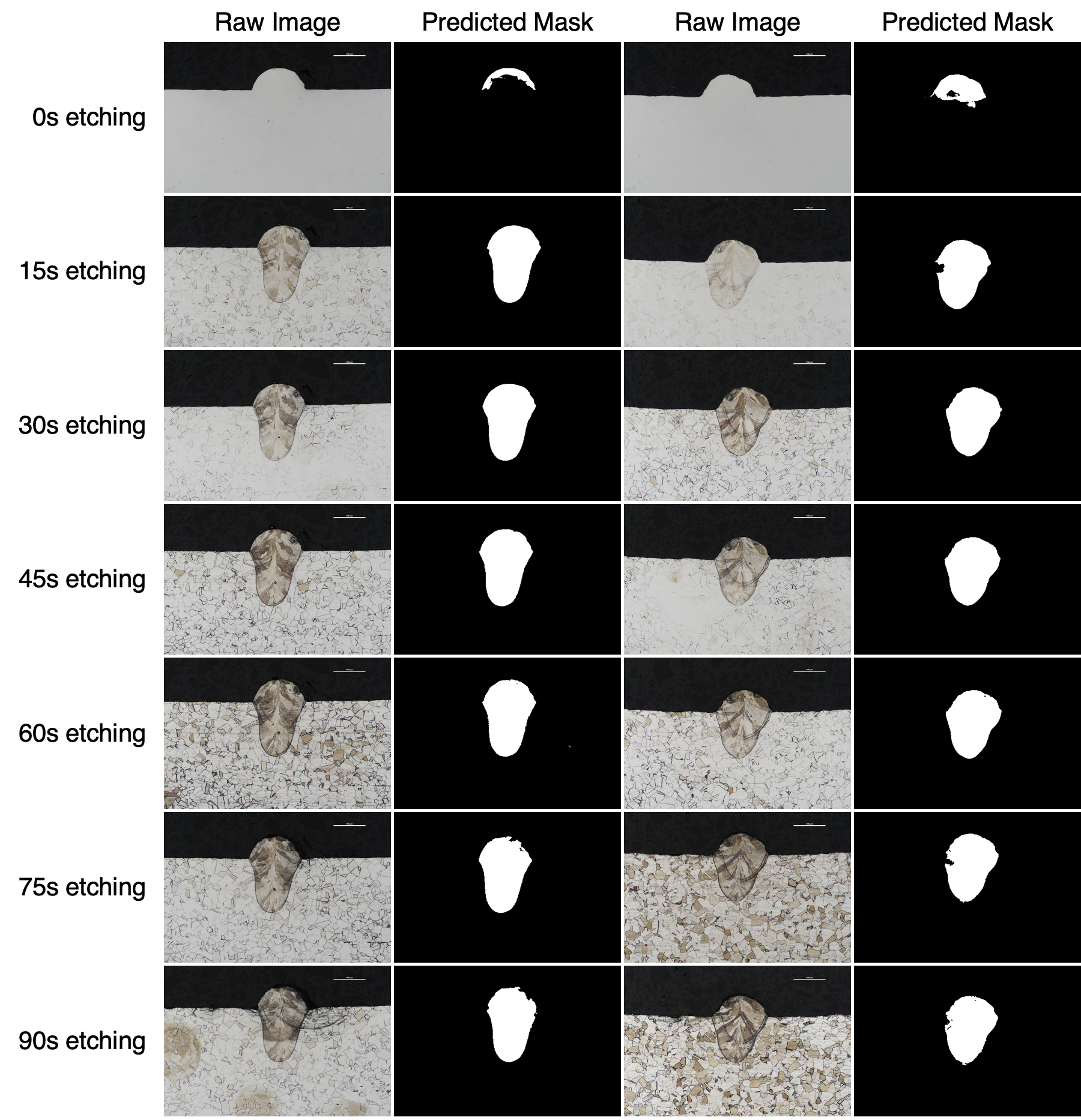}
	\caption{\label{fig:etching-study} Two samples imaged incremental etching with total time progressing from 0s to 90s. The first and second columns show the raw image and predicted mask for the first sample. The third and fourth columns show the same for the second sample. As is visible here, etching is necessary to reveal the melt track boundary, but the model is robust to various contrasts and colour profiles caused by different amounts of etching.}
\end{figure*}

Fig. \ref{fig:etching-study} shows the performance of the model on two samples, both etched from 0 seconds (no etching) to 90 seconds (45 seconds is the normal etching time). Here we observe that the model performs extremely well from 15 seconds onwards. This shows that the model is capable of segmenting a wide range of contrasts introduced by different levels of etching and different experimental protocols. In follow-up work, we have applied the model as trained here to automatically extract melt pool dimensions from a dataset of cross-sections at wide-ranging LPBF parameters, and produce scaling relationships between dimensionless parameters~\cite{weissbach2024workflow}. 

However, the model has some limitations. It occasionally fails to accurately segment the images in the following cases: (1) the image is extremely out of focus; (2) the boundary of the melt pool is visibly hard to discern; (3) the image contains grain boundaries that are similar in contrast and texture to the edge of the melt pool; (4) the image’s color profile markedly differs from the training data (refer to Fig. S1 for exemplary images). Some of these issues can be mitigated by broadening the training data set to include corresponding images, and retraining the model.

\subsection{Potential future directions}

First, the model can be expanded to more rigorously consider defects within melt tracks, such as subsurface pores. Keyhole pores, which are typically circular in cross-section, are some of the most common artefacts in the images. The current model handles keyhole pores differently based on their location. If a pore is entirely within the melt pool, the pore is disregarded and considered as part of the melt pool. However, if the pore lies along the melt pool boundary, it is considered outside the pool. Another printing regime involves ‘balling,’ which is challenging to detect from cross-sectional images. Some telltale signs of balling include wetting angles exceeding 90{\textdegree} and most of the melt pool height being above the surface. The model is capable of accurately segmenting those images as well.  In future work, the model could be expanded to specifically identify and label these defects and calculate metrics such as pore size and position. 

A future direction for the development of this model is its use for multi-track images, namely raster scans for a single layer of LPBF, which have parallel melt tracks with partial overlap. The model is able to accurately segment some images with multiple melt pools, despite being trained on no such images (refer to Fig. S2 for some exemplary images). Addition multi-track images to the training data set would likely improve the accuracy of the model's prediction on such images. 

Given the challenges of generating training data and training a neural network on one method, it also is of interest to apply the current trained model to a different AM process. We tested the model's performance on a few cross-sectional images of samples made by Directed Energy Deposition (DED). The model is able to accurately segment some but not all images (refer to Fig. S2 for some exemplary images). This hints at the possibility of transfer learning~\cite{weiss2016survey}, where our model is used as the starting point for a model geared to DED and trained with a much smaller data set. Separately, a new training set, and trained model could be created for DED or for other AM processes with characteristic image features and dimensions of interest.

\section{Conclusion}

In this study, we demonstrated the use of a neural network based on the U-Net architecture as an effective approach to segment micrographs of the cross-sections of single-track LPBF melt pools. The trained model performs with an accuracy greater than 99\% and an F1 score greater than 90\% despite the small training data set. We show the model's robustness to artefacts, changes in contrast from etching, and changes in colour profile and resolution from different microscopes. We expect the model to be used for future high throughput analysis of LPBF melt pools. We also demonstrate the use of post-processing techniques to collect valuable quantitative metrics from the predicted masks, such as width, height, depth, and wetting and wall angles. We expect that this automated segmentation technique in conjunction with the printing parameters will help expedite the optimisation of high-quality and reliable LPBF. We also highlight some future directions to improve the model and the scope of transfer learning. 

\section{Acknowledgements}

The authors acknowledge support from the National Science Foundation (NSF) through award \#1720701 and \#1922758. This work used Bridges-2 at the Pittsburgh Supercomputing Center through allocation MAT220011 from the Advanced Cyberinfrastructure Coordination Ecosystem: Services \& Support (ACCESS) program, which is supported by National Science Foundation grants \#2138259, \#2138286, \#2138307, \#2137603, and \#2138296. The authors also thank Dr. Brandon Lane, Dr. Lyle Levine, and Dr. Jordan Weaver from the National Institute of Standards and Technology for images used to train the model, along with their guidance. 

\bibliographystyle{elsarticle-num-names}
\bibliography{main.bib}

\clearpage
\onecolumngrid
\renewcommand{\appendixname}{Supplementary Information}
\appendix* 
\renewcommand\thefigure{S\arabic{figure}}    
\section{} 
\setcounter{figure}{0} 

\begin{figure*}[h!]
	\centering
	\includegraphics[width = 0.9\linewidth]{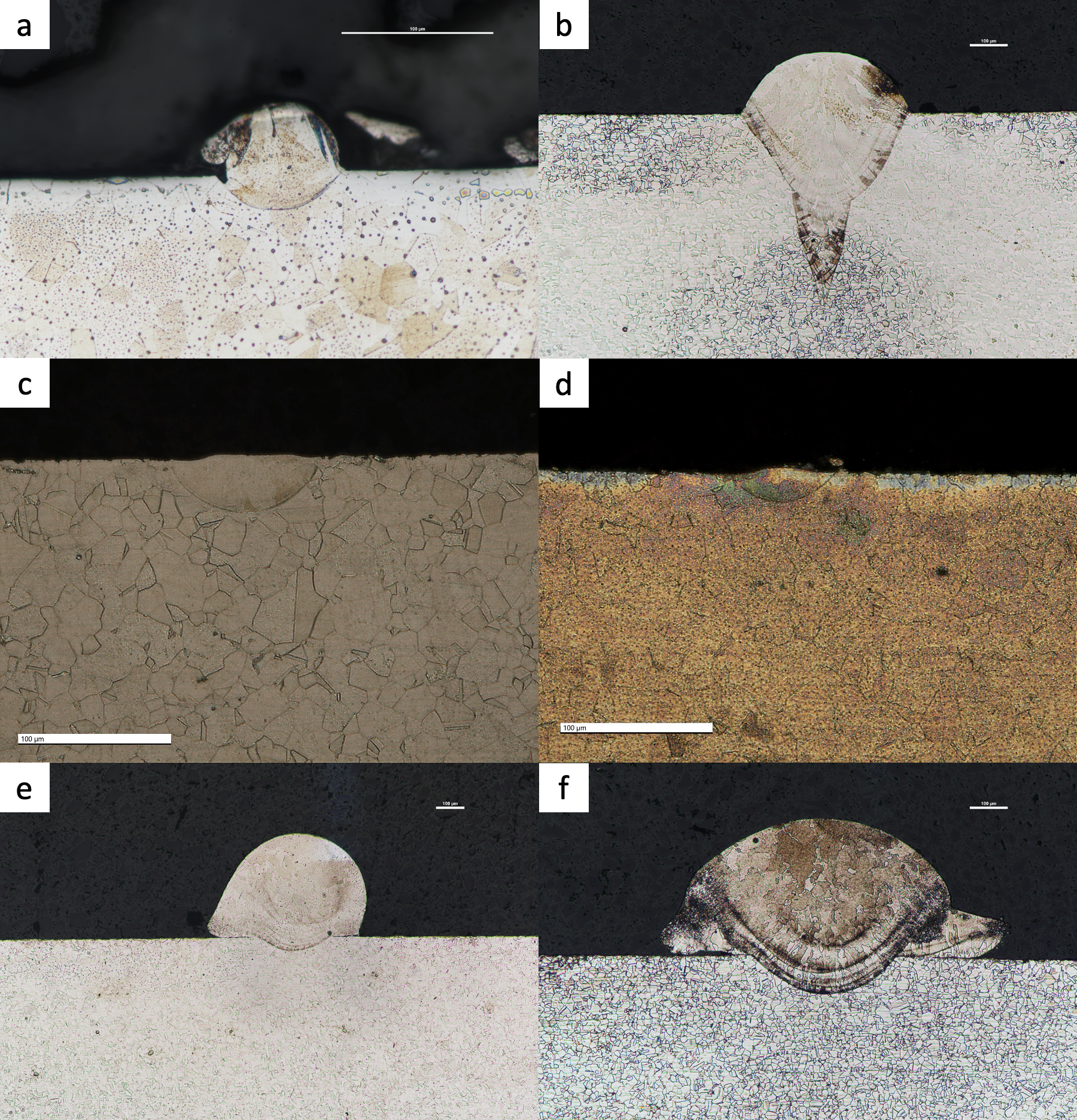}
	\caption{\label{fig:shortcomings} Example images where the model does not perform an accurate segmentation. Each of these images satisfy at least one of these conditions: (1) the image is extremely out of focus, (2) the boundary of the melt pool is visibly hard to discern, (3) there are multiple grain boundaries that are similar to the edge of the melt pool, (4) the image’s color profile markedly differs from the training data.}
\end{figure*}

\begin{figure*}
	\centering
	\includegraphics[width = 0.99\linewidth]{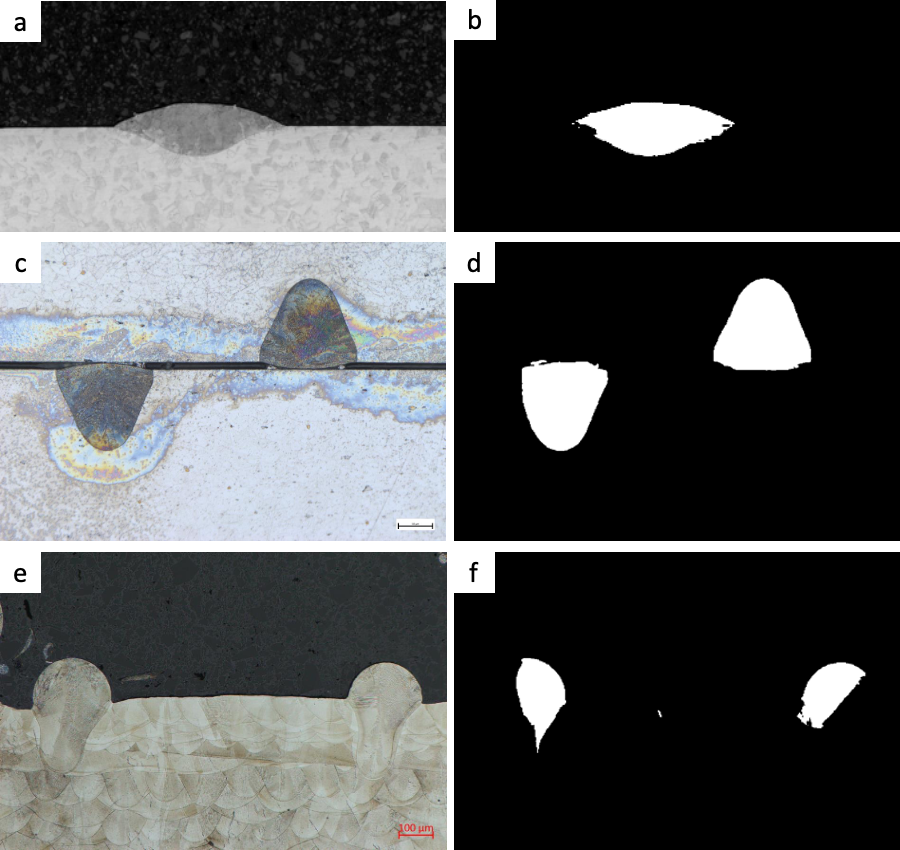}
	\caption{\label{fig:other-images} (a,b) Example image and predicted mask of a cross sectional image of a DED melt track. (c-f) Examples of multi-track cross sectional images and their respective masks. (a) reproduced with permission from~\cite{Jardon2022Effect}}
\end{figure*}

\end{document}